\documentclass[11pt,a4paper]{article}
\PassOptionsToPackage{breaklinks}{hyperref}
\usepackage[hyperref]{acl2020}
\usepackage{times}
\usepackage{latexsym}

\usepackage{microtype}

\usepackage{graphicx}
\usepackage{amssymb}
\usepackage{bm}
\usepackage{caption}
\usepackage{scalefnt}
\usepackage{multirow}
\usepackage{here}
\usepackage{siunitx}
\usepackage{booktabs}
\usepackage{dirtree}
\usepackage{url}
\usepackage{color}
\usepackage[breaklinks]{hyperref}

\aclfinalcopy %
\def\aclpaperid{67} %

\title{ESPnet-ST: All-in-One Speech Translation Toolkit}

\author{Hirofumi Inaguma$^{1}$ ~ Shun Kiyono$^{2}$ ~ Kevin Duh$^{3}$ ~ \textbf{Shigeki Karita}$^{4}$ \\ \textbf{Nelson Yalta}$^5$ ~ \textbf{Tomoki Hayashi}$^{6, 7}$ ~ \textbf{Shinji Watanabe}$^{3}$ \\
${}^{1}$ Kyoto University ~~ ${}^{2}$ RIKEN AIP ~~ ${}^{3}$ Johns Hopkins University \\
${}^{4}$ NTT Communication Science Laboratories ~~ ${}^{5}$ Waseda University \\ ${}^{6}$ Nagoya University ~~ ${}^{7}$ Human Dataware Lab. Co., Ltd. \\
\texttt{inaguma@sap.ist.i.kyoto-u.ac.jp} %
}

\date{}

\begin{document}
\maketitle
\begin{abstract}
We present \textit{ESPnet-ST}, which is designed for the quick development of speech-to-speech translation systems in a single framework.
\textit{ESPnet-ST} is a new project inside end-to-end speech processing toolkit, ESPnet, which integrates or newly implements automatic speech recognition, machine translation, and text-to-speech functions for speech translation.
We provide all-in-one recipes including data pre-processing, feature extraction, training, and decoding pipelines for a wide range of benchmark datasets.
Our reproducible results can match or even outperform the current state-of-the-art performances; these pre-trained models are downloadable.
The toolkit is publicly available at \url{https://github.com/espnet/espnet}.
\end{abstract}

\section{Introduction}
Speech translation (ST), where converting speech signals in a language to text in another language, is a key technique to break the language barrier for human communication.
Traditional ST systems involve cascading automatic speech recognition (ASR), text normalization (e.g., punctuation insertion, case restoration), and machine translation (MT) modules; we call this Cascade-ST~\citep{ney1999speech,casacuberta08,some_insights}.
Recently, sequence-to-sequence (S2S) models have become the method of choice in implementing both the ASR and MT modules (c.f.~\citep{las,attention_nmt_bahdanau}). 
This convergence of models has opened up the possibility of designing end-to-end speech translation (E2E-ST) systems, where a single S2S directly maps speech in a source language to its translation in the target language~\citep{listen_and_translate,weiss2017sequence}.

E2E-ST has several advantages over the cascaded approach: (1) a single E2E-ST model can reduce latency at inference time, which is useful for time-critical use cases like simultaneous interpretation.
(2) A single model enables back-propagation training in an end-to-end fashion, which mitigates the risk of error propagation by cascaded modules.
(3) In certain use cases such as endangered language documentation~\citep{bird2014collecting}, source speech and target text translation (without the intermediate source text transcript) might be easier to obtain, necessitating the adoption of E2E-ST models~\citep{tied_multitask}. 
Nevertheless, the verdict is still out on the comparison of translation quality between E2E-ST and Cascade-ST.
Some empirical results favor E2E~\citep{weiss2017sequence} while others favor Cascade~\citep{iwslt19}; the conclusion also depends on the nuances of the training data condition~\citep{sperber2019attention}.

We believe the time is ripe to develop a unified toolkit that facilitates research in both E2E and cascaded approaches.
We present \textit{ESPnet-ST}, a toolkit that implements many of the recent models for E2E-ST, as well as the ASR and MT modules for Cascade-ST.
Our goal is to provide a toolkit where researchers can easily incorporate and test new ideas under different approaches.
Recent research suggests that pre-training, multi-task learning, and transfer learning are important techniques for achieving improved results for E2E-ST~\citep{audiobook_st,tied_multitask,pretraining_st,inaguma19asru}.
Thus, a unified toolkit that enables researchers to seamlessly mix-and-match different ASR and MT models in training both E2E-ST and Cascade-ST systems would facilitate research in the field.\footnote{There exist many excellent toolkits that support both ASR and MT tasks (see Table \ref{tab:framework}). However, it is not always straightforward to use them for E2E-ST and Cascade-ST, due to incompatible training/inference pipelines in different modules or lack of detailed preprocessing/training scripts. 
}

\begin{table*}[t]
    \centering
    \small
    \tabcolsep 1.2mm
    \begingroup
    \begin{tabular}{l|cccccc|cccccc|c}\toprule
      \multirow{3}{*}{Toolkit} & \multicolumn{6}{c|}{Supported task} & \multicolumn{6}{c|}{\shortstack{Example (w/ corpus pre-processing)}} & \multirow{3}{*}{\shortstack{Pre-trained\\model}} \\ \cline{2-13}
      & \multirow{2}{*}{ASR} & \multirow{2}{*}{LM} & E2E- & Cascade- & \multirow{2}{*}{MT} & \multirow{2}{*}{TTS} & \multirow{2}{*}{ASR} & \multirow{2}{*}{LM} & E2E- & Cascade- & \multirow{2}{*}{MT} & \multirow{2}{*}{TTS} & \\
      & & & ST & ST & & & & & ST & ST & & & \\ \hline
         ESPnet-ST (ours) & \checkmark & \checkmark & \checkmark & \checkmark & \checkmark & \checkmark & \checkmark & \checkmark & \checkmark & \checkmark & \checkmark & \checkmark & \checkmark \\
         Lingvo$^{1}$ & \checkmark & \checkmark & \hspace{2.0mm}$\checkmark^{\clubsuit}$ & \hspace{2.0mm}$\checkmark^{\clubsuit}$ & \checkmark & \hspace{2.0mm}$\checkmark^{\clubsuit}$ & \checkmark & \checkmark & -- & -- & \checkmark & -- & -- \\
         OpenSeq2seq$^{2}$ & \checkmark & \checkmark & -- & -- & \checkmark & \checkmark  & \checkmark & \checkmark & -- & -- & \checkmark & -- & \checkmark \\
         NeMo$^{3}$ & \checkmark & \checkmark & -- & -- & \checkmark & \checkmark  & \checkmark & \checkmark & -- & -- & \checkmark & -- & \checkmark \\
         RETURNN$^{4}$ & \checkmark & \checkmark & \checkmark & -- & \checkmark & -- & -- & -- & -- & -- & -- & -- & \checkmark \\
         SLT.KIT$^{5}$ & \checkmark & -- & \checkmark & \checkmark & \checkmark & -- & \checkmark & -- & \checkmark & \checkmark & \checkmark & -- & \checkmark \\
         Fairseq$^{6}$ & \checkmark & \checkmark & -- & -- & \checkmark & -- & \checkmark & \checkmark & -- & -- & \checkmark & -- & \checkmark \\
         Tensor2Tensor$^{7}$ & \checkmark & \checkmark & -- & -- & \checkmark & -- & -- & -- & -- & -- & \checkmark & -- & \hspace{2.0mm}$\checkmark^\diamondsuit$ \\
         OpenNMT-\{py, tf\}$^{8}$ & \checkmark &  \checkmark & -- & -- & \checkmark & -- & -- & -- & -- & -- & -- & -- & \checkmark \\ 
         Kaldi$^{9}$ & \checkmark & \checkmark & -- & -- & -- & -- & \checkmark & \checkmark & -- & -- & -- & -- & \checkmark \\
         Wav2letter++$^{10}$ & \checkmark & \checkmark & -- & -- & -- & -- & \checkmark & \checkmark & -- & -- & -- & -- & \checkmark \\ 
         \bottomrule
      \end{tabular}
      \caption{Framework comparison on supported tasks in January, 2020. ${}^{\clubsuit}$Not publicly available. ${}^\diamondsuit$Available only in Google Cloud storage.
      $^{1}$\citep{lingvo} $^{2}$\citep{openseq2seq} $^{3}$\citep{nemo} $^{4}$\citep{returnn} $^{5}$\citep{DBLP:journals/pbml/ZenkelSN0PSW18} $^{6}$\citep{fairseq} $^{7}$\citep{tensor2tensor} $^{8}$\citep{opennmt} $^{9}$\citep{kaldi} $^{10}$\citep{wav2letter++}}
      \label{tab:framework}
    \endgroup
\end{table*}

\textit{ESPnet-ST} is especially designed to target the ST task.
ESPnet was originally developed for the ASR task~\citep{espnet}, and recently extended to the text-to-speech (TTS) task~\citep{espnet_tts}.
Here, we extend ESPnet to ST tasks, providing code for building translation systems and  recipes (i.e., scripts that encapsulate the entire training/inference procedure for reproducibility purposes) for a wide range of ST benchmarks.
This is a non-trivial extension: with a unified codebase for ASR/MT/ST and a wide range of recipes, we believe ESPnet-ST is an \textit{all-in-one toolkit} that should make it easier for both ASR and MT researchers to get started in ST research.

The contributions of \textit{ESPnet-ST} are as follows:
\begin{itemize}
    \item To the best of our knowledge, this is the first toolkit to include ASR, MT, TTS, and ST recipes and models in the same codebase.
    Since our codebase is based on the unified framework with a common stage-by-stage processing~\citep{kaldi}, it is very easy to customize training data and models.
    \item We provide recipes for ST corpora such as Fisher-CallHome~\citep{fisher_callhome}, Libri-trans~\citep{libri_trans}, How2~\citep{how2}, and Must-C~\citep{mustc}\footnote{We also support ST-TED~\citep{stted} and low-resourced Mboshi-French~\citep{mboshi_french} recipes.}.
    Each recipe contains a single script (\texttt{run.sh}), which covers all experimental processes, such as corpus preparation, data augmentations, and transfer learning.
    \item We provide the open-sourced toolkit and the pre-trained models whose hyper-parameters are intensively tuned.  
    Moreover, we provide interactive demo of speech-to-speech translation hosted by Google Colab.\footnote{\url{https://colab.research.google.com/github/espnet/notebook/blob/master/st_demo.ipynb}}
\end{itemize}

\begin{figure*}[t]
\begin{minipage}{0.4\hsize}
    \vspace{8mm}
    \centering
    \includegraphics[width=0.80\linewidth]{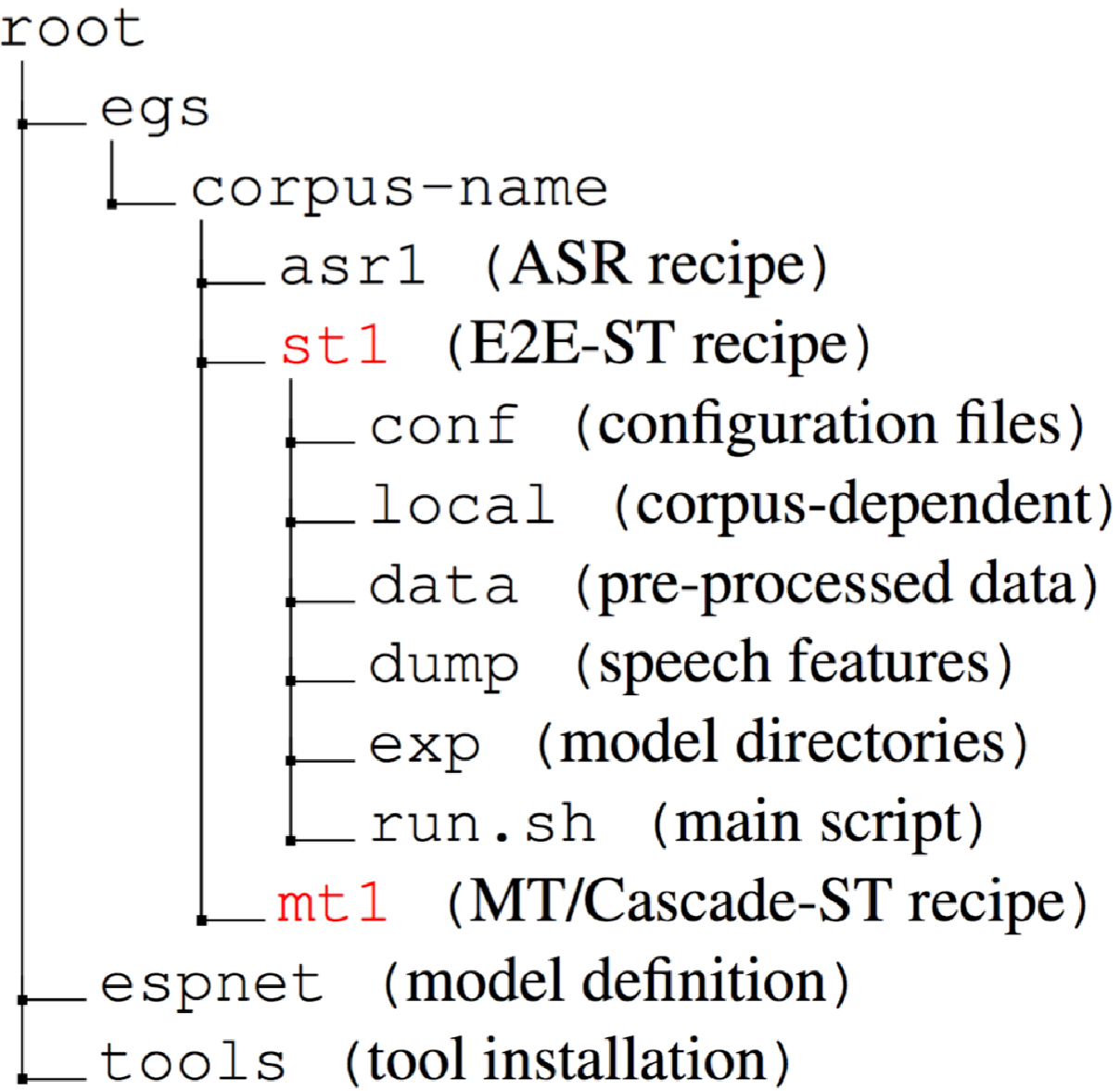}
    \caption{Directory structure of ESPnet-ST}
    \label{fig:directory}
\end{minipage}\hspace{-12mm}
\begin{minipage}{0.75\hsize}
    \centering
    \includegraphics[width=0.78\linewidth]{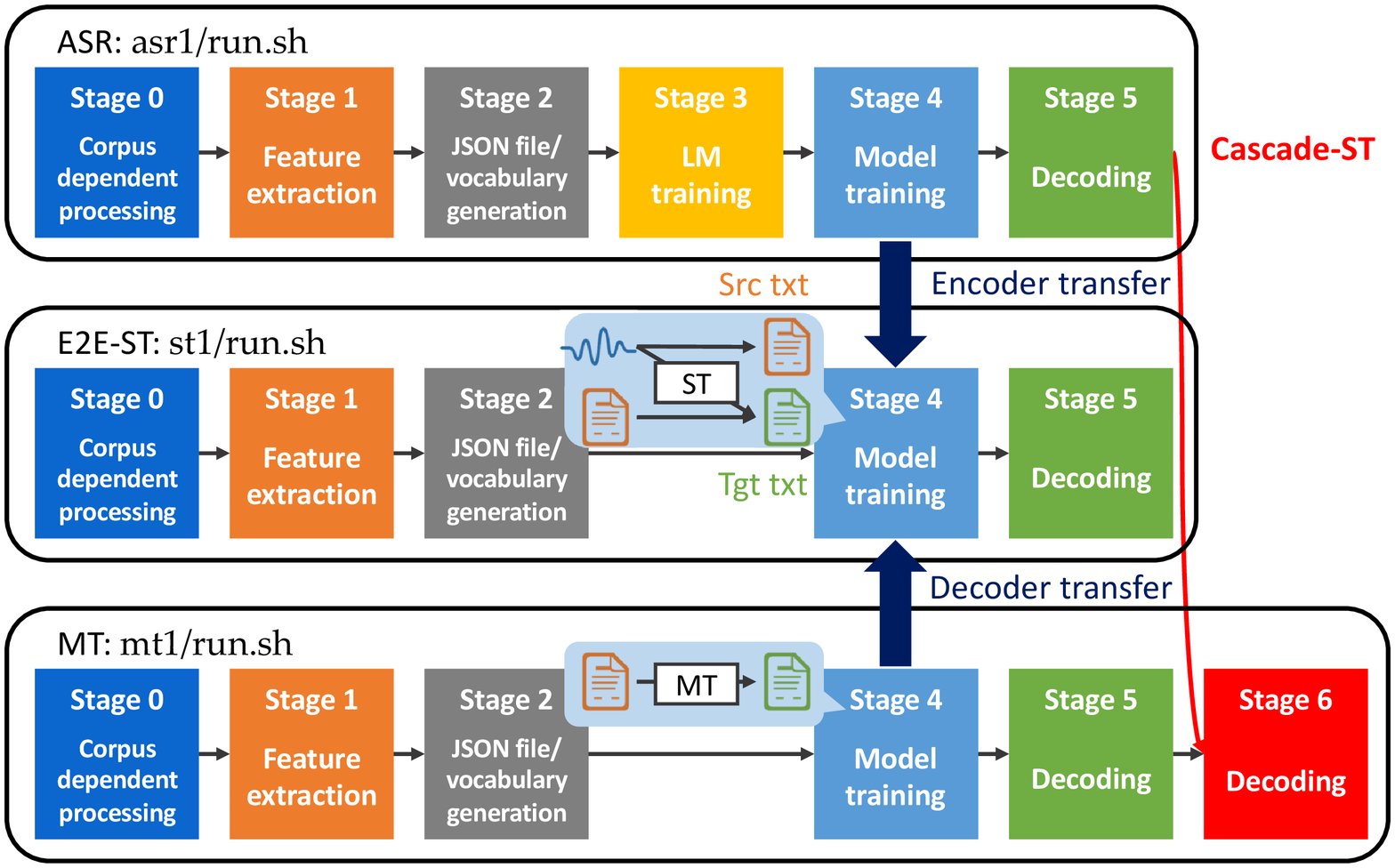}
    \vspace{-1mm}
    \caption{All-in-one process pipelines in ESPnet-ST}
    \label{fig:overview}
\end{minipage}
\end{figure*}

\section{Design}\label{sec:design}

\subsection{Installation}\label{ssec:install}
All required tools are automatically downloaded and built under \texttt{tools} (see Figure \ref{fig:directory}) by a make command.
The tools include (1) neural network libraries such as PyTorch~\citep{pytorch}, (2) ASR-related toolkits such as Kaldi~\citep{kaldi}, and (3) MT-related toolkits such as Moses~\citep{moses} and sentencepiece~\citep{sentencepiece}.
\textit{ESPnet-ST} is implemented with Pytorch backend.

\subsection{Recipes for reproducible experiments}\label{ssec:recipe}
We provide various recipes for all tasks in order to quickly and easily reproduce the strong baseline systems with a single script.
The directory structure is depicted as in Figure \ref{fig:directory}.
\texttt{egs} contains corpus directories, in which the corresponding task directories (e.g., \texttt{st1}) are included.
To run experiments, we simply execute \texttt{run.sh} under the desired task directory.
Configuration yaml files for feature extraction, data augmentation, model training, and decoding etc. are included in \texttt{conf}.
Model directories including checkpoints are saved under \texttt{exp}.
More details are described in Section \ref{ssec:stage}.

\subsection{Tasks}\label{ssec:task}
We support language modeling (LM), neural text-to-speech (TTS) in addition to ASR, ST, and MT tasks.
To the best of our knowledge, none of frameworks support all these tasks in a single toolkit.
A comparison with other frameworks are summarized in Table \ref{tab:framework}.
Conceptually, it is possible to combine ASR and MT modules for Cascade-ST, but few frameworks provide such examples.
Moreover, though some toolkits indeed support speech-to-text tasks, it is not trivial to switch ASR and E2E-ST tasks since E2E-ST requires the auxiliary tasks (ASR/MT objectives) to achieve reasonable performance.

\subsection{Stage-by-stage processing}\label{ssec:stage}
\textit{ESPnet-ST} is based on a stage-by-stage processing including corpus-dependent pre-processing, feature extraction, training, and decoding stages.
We follow Kaldi-style data preparation, which makes it easy to augment speech data by leveraging other data resources prepared in \texttt{egs}.

Once \texttt{run.sh} is executed, the following processes are started.

\vspace{1.5mm}
\noindent\textbf{Stage 0}:
Corpus-dependent pre-processing is conducted using scripts under \texttt{local} and the resulting text data is automatically saved under \texttt{data}.
Both transcriptions and the corresponding translations with three different treatments of casing and punctuation marks (hereafter, punct.) are generated after text normalization and tokenization with \texttt{tokenizer.perl} in Moses; (a) \textbf{tc}: truecased text with punct., (b) \textbf{lc}: lowercased text with punct., and (3) \textbf{lc.rm}: lowercased text without punct. except for apostrophe.
\texttt{lc.rm} is designed for the ASR task since the conventional ASR system does not generate punctuation marks.
However, it is possible to train ASR models so as to generate truecased text using \texttt{tc}.\footnote{We found that this degrades the ASR performance.}

\vspace{1.5mm}
\noindent\textbf{Stage~1}:
Speech feature extraction based on Kaldi and our own implementations is performed.

\vspace{1.5mm}
\noindent\textbf{Stage~2}:
Dataset JSON files in a format ingestable by ESPnet's Pytorch back-end (containing token/utterance/speaker/language IDs, input and output sequence lengths, transcriptions, and translations) are dumped under \texttt{dump}.

\vspace{1.5mm}
\noindent\textbf{Stage~3}:
(ASR recipe only) LM is trained.

\vspace{1.5mm}
\noindent\textbf{Stage~4}:
Model training (RNN/Transformer) is performed.

\vspace{1.5mm}
\noindent\textbf{Stage~5}:
Model averaging, beam search decoding, and score calculation are conducted.

\vspace{1.5mm}
\noindent\textbf{Stage 6}: (Cascade-ST recipe only) The system is evaluated by feeding ASR outputs to the MT model.

\subsection{Multi-task learning and transfer learning}\label{ssec:multitask}
In ST literature, it is acknowledged that the optimization of E2E-ST is more difficult than individually training ASR and MT models.
Multitask training (MTL) and transfer learning from ASR and MT tasks are promising approaches for this problem~\citep{weiss2017sequence,audiobook_st,sperber2019attention,pretraining_st}. 
Thus, in \textbf{Stage 4} of the E2E-ST recipe, we allow options to add auxiliary ASR and MT objectives.
We also support options to initialize the parameters of the ST encoder with a pre-trained ASR encoder in \texttt{asr1}, and to initialize the parameters of the ST decoder with a pre-trained MT decoder in \texttt{mt1}.

\subsection{Speech data augmentation}\label{ssec:data_augmentation}
We implement techniques that have shown to give improved robustness in the ASR component.

\vspace{-1.5mm}
\paragraph{Speed perturbation} We augmented speech data by changing the speed with factors of 0.9, 1.0, and 1.1, which results in 3-fold data augmentation. We found this is important to stabilize E2E-ST training.
\vspace{-7mm}
\paragraph{SpecAugment} Time and frequency masking blocks are randomly applied to log mel-filterbank features. This has been originally proposed to improve the ASR performance and shown to be effective for E2E-ST as well~\citep{bahar2019using}.

\subsection{Multilingual training}
Multilingual training, where datasets from different language pairs are combined to train a single model, 
is a potential way to improve performance of E2E-ST models~\citep{inaguma19asru,di2019one}.
Multilingual E2E-ST/MT models are supported in several recipes.

\subsection{Additional features}\label{sec:implementation}
\paragraph{Experiment manager} We customize the data loader, trainer, and evaluator by overriding Chainer~\citep{tokui2019chainer} modules. The common processes are shared among all tasks.
\vspace{-1.0mm}
\paragraph{Large-scale training/decoding} We support job schedulers (e.g., SLURM, Grid Engine), multiple GPUs and half/mixed-precision training/decoding with \texttt{apex}~\citep{micikevicius2018mixed}.\footnote{\url{https://github.com/NVIDIA/apex}}
Our beam search implementation vectorizes hypotheses for faster decoding~\citep{Seki2019}.
\vspace{-1.0mm}
\paragraph{Performance monitoring} Attention weights and all kinds of training/validation scores and losses for ASR, MT, and ST tasks can be collectively monitored through TensorBoard.
\vspace{-1.0mm}
\paragraph{Ensemble decoding} Averaging posterior probabilities from multiple models during beam search decoding is supported.

\section{Example Models}\label{sec:examplemodel}
To give a flavor of the models that are supported with ESPnet-ST, we describe in detail the construction of an example E2E-ST model, which is used later in the Experiments section. Note that there are many customizable options not mentioned here.

\vspace{-0.5mm}
\paragraph{Automatic speech recognition (ASR)}
We build ASR components with the Transformer-based hybrid CTC/attention framework~\citep{hybrid_ctc_attention}, which has been shown to be more effective than RNN-based models on various speech corpora~\citep{karita2019comparative}.
Decoding with the external LSTM-based LM trained in the \textbf{Stage 3} is also conducted~\citep{shallow_fusion}.
The transformer uses 12 self-attention blocks stacked on the two VGG blocks in the speech encoder and 6 self-attention blocks in the transcription decoder; see~\citep{karita2019comparative} for implementation details.
\vspace{-5.0mm}
\paragraph{Machine translation (MT)}
The MT model consists of the source text encoder and translation decoder, implemented as a transformer with 6 self-attention blocks.
For simplicity, we train the MT model by feeding lowercased source sentences without punctuation marks (\texttt{lc.rm}) \citep{peitz2011modeling}.
There are options to explore characters and different subword units in the MT component.
\paragraph{End-to-end speech translation (E2E-ST)}
Our E2E-ST model is composed of the speech encoder and translation decoder.
Since the definition of parameter names is exactly same as in the ASR and MT components, it is quite easy to copy parameters from the pre-trained models for transfer learning.
After ASR and MT models are trained as described above, their parameters are extracted and used to initialize the E2E-ST model.
The model is then trained on ST data, with the option of incorporating multi-task objectives as well.
\paragraph{Text-to-speech (TTS)}
We also support end-to-end text-to-speech (E2E-TTS), which can be applied after ST outputs a translation.
The E2E-TTS model consists of the feature generation network converting an input text to acoustic features (e.g., log-mel filterbank coefficients) and the vocoder network converting the features to a waveform.
Tacotron~2~\citep{shen2017tacotron2}, Transformer-TTS~\citep{li2018transformer}, FastSpeech~\citep{ren2019fastspeech}, and their variants such as a multi-speaker model are supported as the feature generation network.
WaveNet~\citep{oord2016wavenet} and Parallel WaveGAN~\citep{yamamoto2019parallel} are available as the vocoder network. 
See~\citet{espnet_tts} for more details.

\begin{table*}[t]
    \centering
    \tabcolsep 1mm
    \begingroup
    \small
    \begin{tabular}{c|l|ccccc} \toprule
      \multicolumn{2}{c|}{\multirow{3}{*}{Model}} &  \multicolumn{5}{c}{Es $\to$ En} \\ \cline{3-7}
      \multicolumn{2}{c|}{} & \multicolumn{3}{c}{Fisher} & \multicolumn{2}{c}{CallHome} \\ \cline{3-7}
      \multicolumn{2}{c|}{} & dev & dev2 & test & devtest & evltest \\ \hline
        \multirow{8}{*}{E2E} 
            & Char RNN + ASR-MTL~\citep{weiss2017sequence} & 48.30 & 49.10 & 48.70 & 16.80 & 17.40 \\ \cline{2-7}
            & {\bf ESPnet-ST (Transformer)} & & & & \\
            & ASR-MTL (multi-task w/ ASR)& 46.64 & 47.64 & 46.45 & 16.80 & 16.80 \\
            & \ + MT-MTL (multi-task w/ MT) & 47.17 & 48.20 & 46.99 & 17.51 & 17.64 \\
            & ASR encoder init. (\textcircled{\scriptsize 1}) & 46.25 & 47.11 & 46.21 & 17.35 & 16.94 \\
            & \ + MT decoder init. (\textcircled{\scriptsize 2}) & 46.25 & 47.60 & 46.72 & 17.62 & 17.50 \\
            & \ \ + SpecAugment (\textcircled{\scriptsize 3}) & 48.94 & 49.32 & 48.39 & 18.83 & 18.67 \\
            & \ \ \ + Ensemble 3 models (\textcircled{\scriptsize 1} + \textcircled{\scriptsize 2} + \textcircled{\scriptsize 3}) & {\bf 50.76} & {\bf 52.02} & {\bf 50.85} & {\bf 19.91} & {\bf 19.36} \\ \hline
        \multirow{4}{*}{Cascade} 
            & Char RNN ASR $\to$ Char RNN MT~\citep{weiss2017sequence} & {\bf 45.10} & {\bf 46.10} & {\bf 45.50} & 16.20 & 16.60 \\ 
             & Char RNN ASR $\to$ Char RNN MT~\citep{inaguma19asru}$^{\clubsuit}$ & 37.3 & 39.6 & 38.6 & 16.8 & 16.5 \\ \cline{2-7}
            & {\bf ESPnet-ST} & & & \\
            & Transformer ASR$^{\diamondsuit}$ $\to$ Transformer MT & 41.96 & 43.46 & 42.16 & 19.56 & 19.82 \\ \bottomrule
    \end{tabular}
    \caption{BLEU of ST systems on \underline{Fisher-CallHome Spanish} corpus. ${}^{\clubsuit}$Implemented w/ ESPnet. ${}^\diamondsuit$w/ SpecAugment.}\label{tab:result_fisher_callhome_st}
    \endgroup
\end{table*}

\begin{table}[t]
    \centering
    \tabcolsep 2pt
    \begingroup
    \small
    \begin{tabular}{c|l|c}\toprule
      \multicolumn{2}{c|}{Model} & En $\to$ Fr \\ \hline
        \multirow{12}{*}{E2E} 
            & Transformer + ASR/MT-trans + KD$^{1}$ & 17.02 \\
            & \ + Ensemble 3 models & {\bf 17.8} \\
            & Transformer + PT$^\triangle$ + adaptor$^{2}$ & 16.80 \\ 
            & Transformer + PT$^\triangle$ + SpecAugment$^{3}$ & 17.0 \\ 
            & RNN + TCEN$^{4, \clubsuit}$ & 17.05 \\ \cline{2-3}
            & {\bf ESPnet-ST (Transformer)} &  \\
            & ASR-MTL & 15.30 \\
            & \ + MT-MTL & 15.47 \\
            & ASR encoder init. (\textcircled{\scriptsize 1}) & 15.53 \\
            & \ + MT decoder init. (\textcircled{\scriptsize 2}) & 16.22 \\
            & \ \ + SpecAugment (\textcircled{\scriptsize 3}) & 16.70 \\
            & \ \ \ + Ensemble 3 models (\textcircled{\scriptsize 1} + \textcircled{\scriptsize 2} + \textcircled{\scriptsize 3}) & 17.40 \\ \hline
        \multirow{3}{*}{Cascade} 
            & Transformer ASR $\to$ Transformer MT$^{1}$ & {\bf 17.85} \\ \cline{2-3}
            & {\bf ESPnet-ST} & \\
    & Transformer ASR$^{\diamondsuit}$ $\to$ Transformer MT & 16.96 \\ \bottomrule
    \end{tabular}
    \caption{BLEU of ST systems on \underline{Libri-trans} corpus. $^{\clubsuit}$Implemented w/ ESPnet. ${}^{\triangle}$Pre-training. $^{\diamondsuit}$w/ SpecAugment. $^{1}$\citep{st_distillation} $^{2}$\citep{bahar2019comparative} $^{3}$\citep{bahar2019using} $^{4}$\citep{tcen}}
    \label{tab:result_libri_st}
    \endgroup
\end{table}

\begin{table}[t]
    \centering
    \tabcolsep 1.5pt
    \begingroup
    \small
    \begin{tabular}{c|l|c} \toprule
      \multicolumn{2}{c|}{Model} & En $\to$ Pt \\ \hline
        \multirow{9}{*}{E2E} 
            & RNN~\citep{how2} & 36.0 \\ \cline{2-3}
            & {\bf ESPnet-ST} &  \\
            & Transformer & 40.59 \\
            & \ + ASR-MTL & 44.90 \\
            & \ \ + MT-MTL & 45.10 \\
            & Transformer + ASR encoder init. (\textcircled{\scriptsize 1}) & 45.03 \\
            & \ + MT decoder init. (\textcircled{\scriptsize 2}) & 45.63 \\
            & \ \ + SpecAugment (\textcircled{\scriptsize 3}) & 45.68 \\
            & \ \ \ + Ensemble 3 models (\textcircled{\scriptsize 1} + \textcircled{\scriptsize 2} + \textcircled{\scriptsize 3}) & {\bf 48.04} \\ \hline
        \multirow{2}{*}{Cascade} 
            & {\bf ESPnet-ST} & \\
            & Transformer ASR $\to$ Transformer MT & 44.90 \\ \bottomrule
    \end{tabular}
    \caption{BLEU of ST systems on \underline{How2} corpus}\label{tab:result_how2_st}
    \endgroup
\end{table}

\begin{table*}[t]
    \centering
    \begingroup
    \small
    \begin{tabular}{c|l|cccccccc}\toprule
      \multicolumn{2}{c|}{Model} & De & Pt & Fr & Es & Ro & Ru & Nl & It \\ \hline
        \multirow{4}{*}{E2E} 
            & Transformer + ASR encoder init.$^{1, \clubsuit}$ & 17.30 & 20.10 & 26.90 & 20.80 & 16.50 & 10.50 & 18.80 & 16.80 \\ \cline{2-10}
            & {\bf ESPnet-ST (Transformer)} & & & & & & & \\
            & ASR encoder/MT decoder init. & 22.33 & 27.26 & 31.54 & 27.84 & 20.91 & 15.32 & 26.86 & 22.81 \\
             & \ + SpecAugment & {\bf 22.91} & {\bf 28.01} & {\bf 32.69} & {\bf 27.96} & {\bf 21.90} & {\bf 15.75} & {\bf 27.43} & {\bf 23.75} \\ \hline
        \multirow{3}{*}{Cascade} 
            & Transformer ASR $\to$ Transformer MT$^{1}$ & 18.5 & 21.5 & 27.9 & 22.5 & 16.8 & 11.1 & 22.2 & 18.9 \\ \cline{2-10}
            & {\bf ESPnet-ST} & & & & & & & \\
            & Transformer ASR $\to$ Transformer MT & {\bf 23.65} & {\bf 29.04} & {\bf 33.84} & {\bf 28.68} & {\bf 22.68} & {\bf 16.39} & {\bf 27.91} & {\bf 24.04} \\ \bottomrule
    \end{tabular}
    \caption{BLEU of ST systems on \underline{Must-C} corpus. ${}^{\clubsuit}$Implemented w/ Fairseq. $^{1}$\citep{di2019adapting} }\label{tab:result_mustc_st}
    \endgroup
\end{table*}

\begin{table*}[t]
  \centering
  \small
  \begin{tabular}{ccrrcrr}\toprule
  \multicolumn{1}{l}{} & \multicolumn{3}{c}{En$\to$De} & \multicolumn{3}{c}{De$\to$En}  \\ 
  \cmidrule(r){2-4} \cmidrule(r){5-7}
  Framework & test2012 & \multicolumn{1}{c}{test2013} & \multicolumn{1}{c}{test2014} & test2012 & \multicolumn{1}{c}{test2013} & \multicolumn{1}{c}{test2014} \\\midrule
  Fairseq & \multicolumn{1}{r}{27.73} & 29.45 & 25.14   & \multicolumn{1}{r}{32.25} & 34.23 & 29.49 \\
  ESPnet-ST & \multicolumn{1}{r}{26.92} & 28.88 & 24.70 & \multicolumn{1}{r}{32.19} & 33.46 & 29.22 \\
  \bottomrule
  \end{tabular}
  \caption{BLEU of MT systems on \underline{IWSLT 2016} corpus}\label{tab:result_iwslt16}
\end{table*}

\section{Experiments}
In this section, we demonstrate how models from our ESPnet recipes perform on benchmark speech translation corpora: Fisher-CallHome Spanish En$\to$Es, Libri-trans En$\to$Fr, How2 En$\to$Pt, and Must-C En$\to$8 languages.
Moreover, we also performed experiments on IWSLT16 En-De to validate the performance of our MT modules.

All sentences were tokenized with the \texttt{tokenizer.perl} script in the Moses toolkit~\citep{moses}.
We used the joint source and target vocabularies based on byte pair encoding (BPE)~\citep{sennrich2015neural} units. 
ASR vocabularies were created with English sentences only with \texttt{lc.rm}.
We report 4-gram BLEU~\citep{bleu} scores with the \texttt{multi-bleu.perl} script in Moses.
For speech features, we extracted 80-channel log-mel filterbank coefficients with 3-dimensional pitch features using Kaldi, resulting 83-dimensional features per frame.
Detailed training and decoding configurations are available in \texttt{conf/train.yaml} and \texttt{conf/decode.yaml}, respectively.

\subsection{Fisher-CallHome Spanish (Es$\to$En)}
Fisher-CallHome Spanish corpus contains 170-hours of Spanish conversational telephone speech, the corresponding transcription, as well as the English translations~\citep{fisher_callhome}.
All punctuation marks except for apostrophe were removed~\citep{fisher_callhome,some_insights,weiss2017sequence}.
We report case-insensitive BLEU on Fisher-\{{\em dev}, {\em dev2}, {\em test}\} (with four references), and CallHome-\{{\em devtest}, {\em evltest}\} (with a single reference).
We used 1k vocabulary for all tasks.

Results are shown in Table \ref{tab:result_fisher_callhome_st}.
It is worth noting that we did not use any additional data resource.
Both MTL and transfer learning improved the performance of vanilla Transformer.
Our best system with SpecAugment matches the current state-of-the-art performance~\citep{weiss2017sequence}.
Moreover, the total training/inference time is much shorter since our E2E-ST models are based on the BPE1k unit rather than characters.\footnote{\citet{weiss2017sequence} trained their model for more than 2.5 weeks with 16 GPUs, while \textit{ESPnet-ST} requires just 1-2 days with a single GPU. The fast inference of ESPnet-ST can be confirmed in our interactive demo page (RTF 0.7755).}

\subsection{Libri-trans (En$\to$ Fr)}
Libri-trans corpus contains 236-hours of English read speech, the corresponding transcription, and the French translations~\citep{libri_trans}.
We used the clean 100-hours of speech data and augmented translation references with Google Translate for the training set~\citep{audiobook_st,st_distillation,bahar2019comparative,bahar2019using}.
We report case-insensitive BLEU on the {\em test} set.
We used 1k vocabulary for all tasks.

Results are shown in Table \ref{tab:result_libri_st}.
Note that all models used the same data resource and are competitive to previous work.

\subsection{How2 (En$\to$ Pt)}
How2 corpus contains English speech extracted from YouTube videos, the corresponding transcription, as well as the Portuguese translation~\citep{how2}.
We used the official 300-hour subset for training.
Since speech features in the How2 corpus is pre-processed as 40-channel log-mel filterbank coefficients with 3-dimensional pitch features with Kaldi in advance, we used them without speed perturbation.
We used 5k and 8k vocabularies for ASR and E2E-ST/MT models, respectively.
We report case-sensitive BLEU on the {\em dev5} set.

Results are shown in Table \ref{tab:result_how2_st}.
Our systems significantly outperform the previous RNN-based model~\citep{how2}.
We believe that our systems can be regarded as the reliable baselines for future research.

\subsection{Must-C (En$\to$ 8 langs)}
Must-C corpus contains English speech extracted from TED talks, the corresponding transcription, and the target translations in 8 language directions (De, Pt, Fr, Es, Ro, Ru, Nl, and It)~\citep{mustc}. 
We conducted experiments in all 8 directions.
We used 5k and 8k vocabularies for ASR and E2E-ST/MT models, respectively.
We report case-sensitive BLEU on the {\em tst-COMMON} set.

Results are shown in Table \ref{tab:result_mustc_st}.
Our systems outperformed the previous work \citep{di2019adapting} implemented with the custermized Fairseq\footnote{\url{https://github.com/mattiadg/FBK-Fairseq-ST}} with a large margin.

\subsection{MT experiment: IWSLT16 En $\leftrightarrow$ De}
IWSLT evaluation campaign dataset~\citep{cettolo2012wit3} is the origin of the dataset for our MT experiments.
We used En-De language pair.
Specifically, IWSLT 2016 training set for training data, test2012 as the development data, and test2013 and test2014 sets as our test data respectively.

We compare the performance of Transformer model in \textit{ESPnet-ST} with that of Fairseq in Table~\ref{tab:result_iwslt16}.
\textit{ESPnet-ST} achieves the performance almost comparable to the Fairseq.
We assume that the performance gap is due to the minor difference in the implementation of two frameworks.
Also, we carefully tuned the hyper-parameters for the MT task in the small ST corpora, which is confirmed from the reasonable performances of our Cascaded-ST systems.
It is acknowledged that Transformer model is extremely sensitive to the hyper-parameters such as the learning rate and the number of warmup steps~\citep{popel2018training}.
Thus, it is possible that the suitable sets of hyper-parameters are different across frameworks.

\section{Conclusion}
We presented \textit{ESPnet-ST} for the fast development of end-to-end and cascaded ST systems.
We provide various all-in-one example scripts containing corpus-dependent pre-processing, feature extraction, training, and inference.
In the future, we will support more corpora and implement novel techniques to bridge the gap between end-to-end and cascaded approaches.

\section*{Acknowledgment}
We thank Jun Suzuki for providing helpful feedback for the paper.

\bibliography{reference}
\bibliographystyle{acl_natbib}

\end{document}